# TPC-ViT: Token Propagation Controller for Efficient Vision Transformers

Wentao Zhu

## Abstract

*Vision transformers (ViTs) have achieved promising results on a variety of Computer Vision tasks, however their quadratic complexity in the number of input tokens has limited their application specially in resource-constrained settings. Previous approaches that employ gradual token reduction to address this challenge assume that token redundancy in one layer implies redundancy in all the following layers. We empirically demonstrate that this assumption is often not correct, i.e., tokens that are redundant in one layer can be useful in later layers. We employ this key insight to propose a novel token propagation controller (TPC) that incorporates two different token-distributions, i.e., pause probability and restart probability to control the reduction and reuse of tokens respectively, which results in more efficient token utilization. To improve the estimates of token-distributions, we propose a smoothing mechanism that acts as a regularizer and helps remove noisy outliers. Furthermore, to improve the training-stability of our proposed TPC, we introduce a model stabilizer that is able to implicitly encode local image structures and minimize accuracy fluctuations during model training. We present extensive experimental results on the ImageNet-1K dataset using DeiT, LV-ViT and Swin models to demonstrate the effectiveness of our proposed method. For example, compared to baseline models, our proposed method improves the inference speed of the DeiT-S by 250% while increasing the classification accuracy by 1.0%.*

## 1. Introduction

Vision transformers (ViTs) [6,19,31] have achieved impressive results on a wide variety of tasks, including semantic segmentation [16, 35], image generation [13, 14], object detection [2, 3, 27], and image classification [6, 11, 19, 31, 34, 40]. Despite their initial success, the quadratic computational complexity of ViTs to the number of input tokens [28,38] continues to be a key challenge that limits their application particularly in resource-constrained settings.

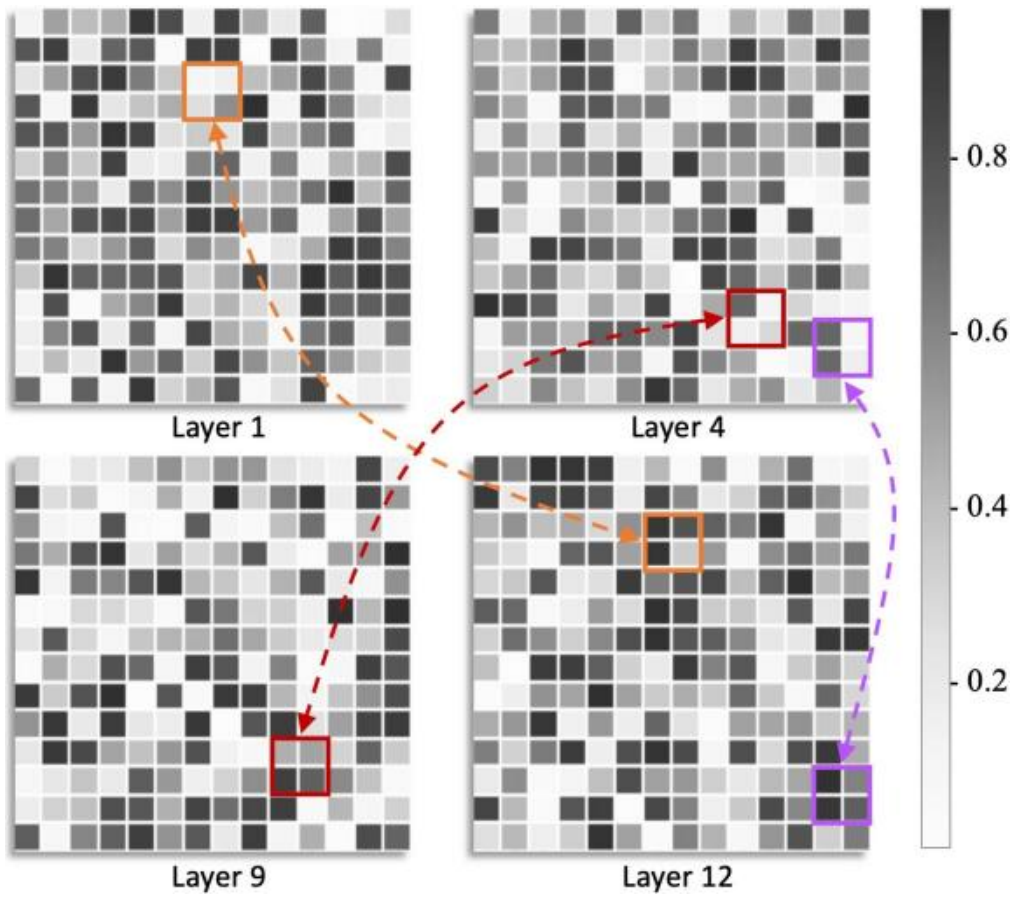

Figure 1. Visualization of token attention maps based on CLS tokens in ViT. The attention weights for each token are not always large or small in different layers, *e.g.*, the tokens annotated with orange, red, and purple square boxes. TPC-ViT decouples the break probability with a pause probability and a restart probability to mitigate this challenge.

An important direction explored to address this challenge is to gradually reduce the number of tokens based on various selection criteria [17, 23, 39]. However, this class of methods assumes that a token considered to be redundant in a particular transformer-layer is also redundant in all the following layers. That is, once a token is removed, it is discarded forever. However, we demonstrate in Figure 1 that this assumption is often inaccurate, and tokens that are redundant in one layer can often be useful in later layers.

Building on this key observation, we propose a novel token propagation controller (TPC) that incorporates two different random variables, *i.e.*, pause probability and restart probability for each token, such that the product of these two probabilities controls whether a token gets to be removed or reused across different transformer layers (see Figure 2 and Figure 3 for details). We identify that accurate estimation of pause and restart token probabilities can be challenging in practice due to factors such as image noise, motion or lens blur, and small foreground regions. We ad-dress this challenge by proposing a distribution regularizer that normalizes the pause and restart probabilities of each token using the mean pause and restart probabilities computed over all the tokens in an image. These mean pause and restart probabilities act as global priors by encoding the overall characteristics of an image, and enable us to effectively adjust each individual token's local pause and reuse probability estimates.

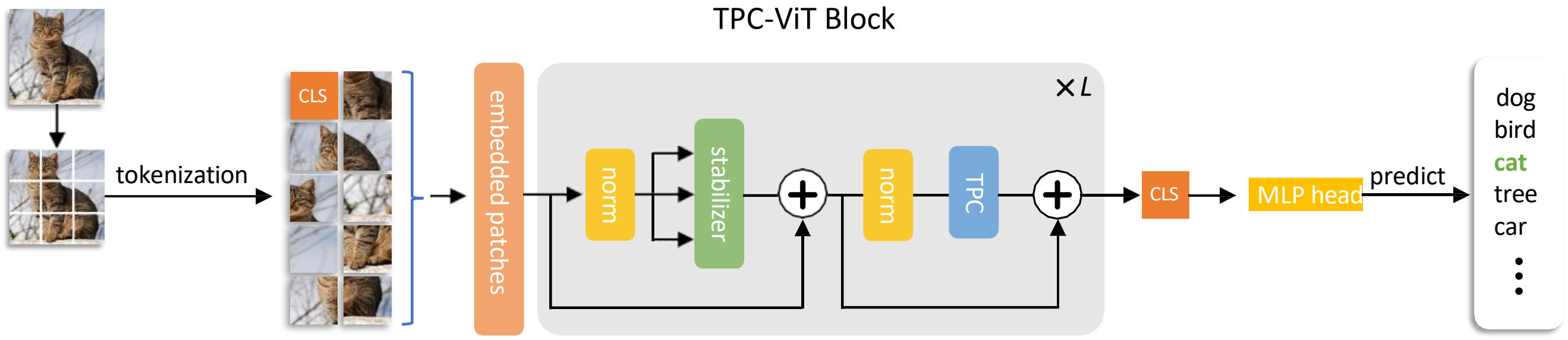

Figure 2. Framework of TPC-ViT, where each block is mainly composed of a stabilizer and a token propagation controller (TPC). The purpose of the stabilizer is to improve the convergence speed of the proposed joint probability-based method by using a specific amount of similar tokens.

Furthermore, we identify that training our proposed TPC can show accuracy fluctuations most likely due to the dense attention map calculation without incorporating any local neighborhood bias. We address this challenge by introducing a model-stabilizer that uses similar tokens of the keys for each query for sparse attention map computation. As similar tokens are more likely to be found within the local neighborhood of query tokens, our proposed stabilizer can naturally incorporate local neighborhood bias (as offered by CNNs) without actually using any convolutions [33, 41, 42, 43].

These novel design choices enable us to train our TPC in a stable manner and adaptively determine the tokens that need to be removed or reused across various transformer layers. We present extensive empirical comparisons of our approach with DeiT, LV-ViT and Swin models on ImageNet-1K dataset, which show that our method achieves both inference speed-up as well as performance increase. Specifically, our method improves the inference speed of DeiT-S by $250\%$ while increasing the classification accuracy by $1.0\%$.

The **key contributions** of our work are summarized below:

- A novel token propagation approach that adaptively removes and reuses tokens in different transformer layers, resulting in overall more effective token utilization.
- A simple yet effective token distribution regularizer that acts as a global prior by encoding the overall characteristics of an image and enables effective adjustment of individual token's probability estimates.
- A novel model-stabilizer that naturally incorporates local neighborhood bias and helps minimize accuracy fluctuations particularly during the early part of training.
- Extensive comparative experiments showing the effectiveness of our proposed approach to both accelerate the inference speed as well as increase the model accuracy.

## 2. Related Work

In addition to being deeply explored in Natural Language Processing [30], previous works in Computer Vision have looked at efficient transformers mostly from two main perspectives including: (a) efficient self-attention, and (b) token reduction. In the following we go over some of the main works in these two directions.

### 2.1. Efficient Attention Mechanisms

To reduce the quadratic complexity of transformer-based models with respect to input number of tokens [32], various efficient attention mechanisms have been explored in the past. For instance, Swin Transformer [19] introduced a shifted windowing scheme by limiting self-attention computation to non-overlapping local windows while also allowing for cross-window connections. CoaT [36] factorized the attention and calculated the self-attention between keys and values first. In X-ViT [25], linear self-attention is developed by replacing softmax with Xnorm. Similarly, in SOFT [20], a Gaussian kernel function is adopted to replace the dot-product similarity without softmax normalization. DaViT [5] improved the efficiency by grouping tokens both in spatial and channel dimensions. Unlike these works that mainly focus on efficient attention mechanisms to reduce transformer complexity, our focus is on the direction of token reduction for improving transformer efficiency.

### 2.2. Token Reduction Approaches

Given the considerable redundancy often found in the input image tokens, removing tokens based on various selection criteria is another main direction that has been explored in the past to accelerate transformers [8, 17, 23, 39]. For instance, Dynamic-ViT [23] proposed a dynamic token sparsification framework to prune redundant tokens progressively and dynamically based on a particular input. A lightweight prediction module added to different layers to hierarchically prune redundant tokens is introduced to estimate the importance score of each token given the current features. EViT [17] introduced a module to reorganize image tokens during the feed-forward process of vision transformers. For each forward inference, the attentive tokens between the multi-head self-attention and feed-forward network modules are identified based on the corresponding class token attention. Similarly, AViT [39] proposed a halting mechanism to adaptively discard redundant image tokens. This method is capable of effectively reducing the number of tokens in vision transformers that are processed in the network as inference proceeds. Note that the gradual token reduction methods [17, 23, 39] in this direction assume that once a token is considered to be redundant in one layer, it is removed from contributing in any of the following layers. We demonstrate that this assumption is often not correct, and propose a novel token propagation mechanism that can adaptively remove or reuse tokens in different layers to achieve more effective token utilization. Our comparative results demonstrate that our proposed method outperforms previous state-of-the-art token reduction approaches by a large margin.

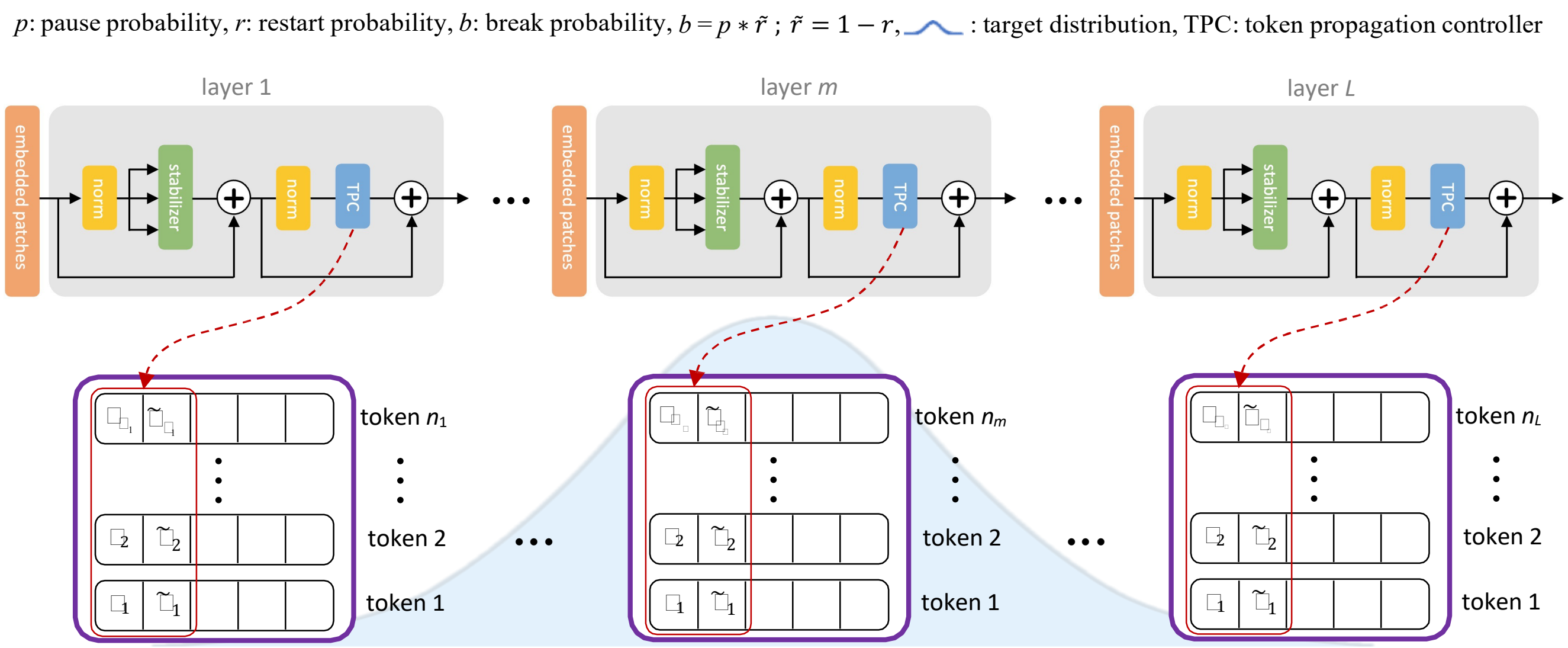

Figure 3. Illustration of our proposed joint probability-based TPC. The total number of layers is denoted by $L$, $m$ indicates a middle layer index, $n_i$ denotes the total number of tokens in the layer $i$, and the purple round rectangle denotes the introduced token distribution regularizer. The break probability $b$ can be factorized into a joint probability of the pause probability $p$ and the restart probability $\tilde{r}$. The light blue bell shape distribution denotes a target distribution across layers. If the propagation of a token is too deep, the computational cost is high, *i.e.*, inefficient. However, if the propagation stops too early, the information loss is large. Hence, The target distribution, *e.g.*, Gaussian, is used to implicitly encourage tokens to terminate at the expected target layer $M_k$. Please see Equation (6), (12), and (13) for more details.

# 3. Approach

We now present the details of our proposed method where § 3.1 explains how we adaptively control the propagation of tokens, while § 3.2 explains how we use a distribution regularizer for each individual token. The derivation of our training objective is given in§ 3.3, while §3.4 shows how we incorporate our proposed stabilizer into our TPC-ViT to improve its training convergence speed.

## 3.1. Token Propagation Controller

We formulate toke propagation probability in vision transformers as a joint probability and factorize it into two parts, *i.e.*, (a) a pause probability $p$ and (b) a restart prob- ability $r$. Pause probability determines that the token is temporally not used in a particular layer, while restart probability indicates the chance that a token might be used in any of the following layers. Two random variables X and Y are introduced to characterize their behaviors and we model their joint probability mass function as conditional independent variables. A break probability $b$ is introduced to determine whether the propagation of a token should terminate or not. The break probability $b$ is defined as the multiplication of the pause probability $p$ and the non-restart probability $\tilde{r} = 1 - r$. The visual illustration is shown in Figure 3.

**Definition 1.** Let X and Y be two binary random variables with Bernoulli distributions respectively. The joint probability mass function of (X, Y) is then defined as:

$$\text{prob}(p, \tilde{r}) = \text{Prob}(\text{X} = p, \text{Y} = \tilde{r}) \tag{1}$$

Note that total number of tokens is fixed from one layer to another in the computation flow of vanilla ViT blocks, *i.e.*,

$$t^{l}_{\{i\in \mathbb{N}|i\in\{1,2,\dots,K\}\}} = \text{ViT}^{l}_{\text{vanilla}}\left(t^{l-1}_{\{j\in \mathbb{N}|j\in\{1,2,\dots,K\}\}}\right) \quad (2)$$

where $K$ denotes the total number of tokens, $\text{ViT}^{l}_{\text{vanilla}}(\cdot)$ denotes the vanilla ViT block at layer $l$, $t^{l}_{\{i\in \mathbb{N}|i\in\{1,2,\dots,K\}\}}$ indicates all the $K$ updated tokens, and $t^{l-1}_{\{j\in \mathbb{N}|j\in\{1,2,\dots,K\}\}}$ denotes the $K$ tokens before the updated.

In our proposed TPC-ViT, the total number of tokens can be changed from one layer to another, as shown in the computation flow in Figure 3, *i.e.*,

$$t^{l}_{\{i\in \mathbb{N}|i\in S\subseteq\{1,2,\dots,K\}\}} = \text{TPC}^{l}_{\text{vit}}\left(t^{l-1}_{\{j\in \mathbb{N}|j\in\{1,2,\dots,K\}\}}\right) \quad (3)$$

where $\text{TPC}^{l}_{\text{vit}}(\cdot)$ denotes our proposed TPC-ViT block at the layer $l$, $t^{l}_{\{i\in \mathbb{N}|i\in S\subseteq\{1,2,\dots,K\}\}}$ indicates all the $|S|$ updated tokens, while $t^{l-1}_{\{j\in \mathbb{N}|j\in\{1,2,\dots,K\}\}}$ indicates all the $K$ tokens before updated. Note that S is a subset of $\{1, 2, ..., K\}$ and $|S|$ denotes the total number of elements in S. The size of S is learnable, *i.e.*, dynamically changing in each layer. See Figure 3 for more details.

To adaptively terminate one token, the break probability of the $k^{\text{th}}$ token at the layer $l$ can then be calculated as:

$$\begin{aligned} b^l_k &= \text{prob}(p^l_k, r^{\sim l}_k) = \text{Prob}(X = p^l_k, Y = r^{\sim l}_k) \\ &= \text{Prob}(X = p^l_k)\cdot\text{Prob}(Y = r^{\sim l}_k|X) = P_1(t^l_k)\cdot P_2(t^l_k) \end{aligned} \quad (4)$$

where $P_1(\cdot)$ and $P_2(\cdot)$ are used to compute the pause and non-restart probabilities respectively. The two probabilities are incorporated into existing vision transformer blocks by reusing two existing neurons in the multi-layer perceptron layer, *i.e.*, the blue block and purple round rectangle in Figure 3. Hence there are no additional learnable parameters introduced for our proposed TPC-ViT except for the two scalar parameters $\gamma$ and $\beta$ in Equation (5). The two scalar parameters are used to tune the break probability of each token $b^l_k$ to be in the range $0 \le b^l_k \le 1$, *i.e.*,

$$P_1(t^l_k) = \frac{1}{1+e^{-(\gamma\cdot t^l_{k,1}+\beta)}}, \quad P_2(t^l_k) = \frac{1}{1+e^{-(\gamma\cdot t^l_{k,2}+\beta)}} \quad (5)$$

where $t^l_{k,1}$ and $t^l_{k,2}$ indicate the first and second embedding dimensions of the token $t^l_k$. We empirically find that using the first two dimensions gives us sufficiently good results, and varying positions of the dimensions does not affect the model's performance much (see Table 1 for more details). The shifting parameter $\gamma$ and the scaling parameter $\beta$ are shared for all tokens across all layers.

In this work, the cumulative break probability is used to terminate tokens as inference progresses into deeper layers. When the cumulative break probability exceeds $1-\delta$, the proposed model will conduct the token reduction at the $M_k$ layer, *i.e.*,

$$M_k = \underset{n\le L}{\arg\min} \sum_{l=1}^{n} b^l_k = \arg\min \sum_{l=1}^{n} P_1(t^l_k)P_2(t^l_k) \ge 1-\delta \quad (6)$$

where $\delta$ denotes an arbitrarily small positive number used for token reduction after one layer, and $L$ indicates the total number of TPC-ViT blocks or layers.

| Model | Top-1 | Top-5 | Params (M) |
|---|---|---|---|
| 2 RVs (3rd, 6th) | 79.9 | 95.3 | 22 |
| 2 RVs (7th, 8th) | 80.0 | 95.5 | 22 |
| **2 RVs (1st, 2nd)** | **80.1** | **95.7** | **22** |

Table 1. Varying dimension positions for random variables (RVs) used for break probability based on proposed TPC-DeiT-S with $\kappa$ value of 100.

To facilitate the calculation of Equation (6), a remainder $\hat{r}$ for each token is defined to track the progress of break probabilities across layers as:

$$\hat{r}_k = 1 - \sum_{l=1}^{M_k-1} b^l_k = 1 - \sum_{l=1}^{M_k-1} P_1(t^l_k)P_2(t^l_k) \quad (7)$$

After we aggregate Equation (4), (5), (6) and (7), the final aggregated break probability considering the token reduction conducted layer $M_k$ can be formalized as:

$$p^l_k = \begin{cases} 0, & \text{if } l > M_k \\ \hat{r}_k = 1 - \sum_{l=1}^{M_k-1} P_1(t^l_k)P_2(t^l_k), & \text{if } l = M_k \\ b^l_k = P_1(t^l_k)P_2(t^l_k), & \text{if } l < M_k \end{cases} \quad (8)$$

where $p^l_k \in [0,1] = \{x \in \mathbb{R}|0 \le x \le 1\}$. This will be used for the final output token calculation in the next section.

### 3.2. Token Distribution Regularizer

Accurate estimation of pause and restart token probabilities can be challenging in practice due to factors such as image noise, motion or lens blur, and small foreground regions. To address this challenge, we propose a token distribution regularizer that normalizes the pause and restart probabilities of each token using the average pause and restart probabilities computed over all the tokens in an image. These average probabilities act as global priors by encoding the overall characteristics of an image, and enable effective adjustment of each token's pause and reuse probability estimates. The regularized break probability is formalized as:

$$b^l_{k_{\text{regularized}}} = \xi\cdot b^l_k + (1-\xi)\cdot\frac{1}{|S|}\sum_{k=1}^{|S|} b^l_k \quad (9)$$

where $\frac{1}{|S|}\sum_{k=1}^{|S|} b_k^l$ is the introduced token distribution regularizer.

### 3.3. Training Objective

The training objective function for our proposed TPC-ViT method is composed of three parts, *i.e.*, (a) a ponder loss $L_{\text{ponder}}$, (b) a task loss $L_{\text{task}}$, and (c) a distribution loss $L_{\text{distribution}}$. Since deeper token propagation results higher overall computational cost, the purpose of $L_{\text{ponder}}$ is to encourage early stopping. The overall ponder loss is formulated via the auxiliary variable $\hat{r}$ as:

$$L_{\text{ponder}} := \frac{1}{|S|}\sum_{k=1}^{|S|}(M_k + \hat{r}_k), \tag{10}$$

where S is the token index set. The ponder loss averages the combination of terminated layer number and remainder across all the tokens.

Similar to other vision transformer models, we use the CLS token $t_{\text{cls}}$ to yield the classification prediction in our TPC-ViT, and it is updated in all layers together with other tokens. The output token $t_{\text{output}}$ is calculated based on the aggregated break probability weighted average over previous states (see Equation (8) for the definition of aggregated break probability $p_k^l$). The associated task loss $L_{\text{task}}$ is defined as:

$$L_{\text{task}} := \text{Post}(t_{\text{output}}) = \text{Post}(\sum_{l=1} p_{\text{cls}}^l t_{\text{cls}}^l), \tag{11}$$

where Post(·) denotes a post-processor for the transformed CLS token after the entire stack, which can be a cross-entropy loss function for image classification.

At this stage, a temporary loss can be defined to explain the training procedure of the proposed TPC-ViT based on the $L_{\text{task}}$ and the $L_{\text{ponder}}$ with a scaling factor $\phi_p$ in Eqution (14). The factor $\phi_p$ is used to scale the pondering loss $L_{\text{ponder}}$ relative to the task loss $L_{\text{task}}$. When a larger value of $\phi_p$ is used in the training phase, it imposes a stronger penalty over token termination layer number. Hence, the model learns to halt tokens earlier. Although $\phi_p$ is capable of balancing the strength of token reduction and network performance for the target application, the choice of $\phi_p$ can be sensitive to the training of the adaptive computation-based vision transformer models [9, 12]. The value of $\phi_p$ may not provide a sufficiently fine-grained control over the trade-off between accuracy and efficiency.

One way to address the challenge is that, in addition to $\phi_p$, we also introduce a prior distribution to regularize the layer-level break probability $D$, such that tokens are expected to stop at a target depth on average, but we still allow variations for each image. As described in [1], the prior distribution helps stabilize convergence during stochastic pondering. To this end, a layer-level break probability distribution $D \in \mathbb{R}^L$ is defined based on averaging the break probability for all tokens across each layer of the network as:

$$D := \frac{1}{|S|}\sum_{k=1}^{|S|}[b_k^1, b_k^2, \cdots, b_k^L]. \tag{12}$$

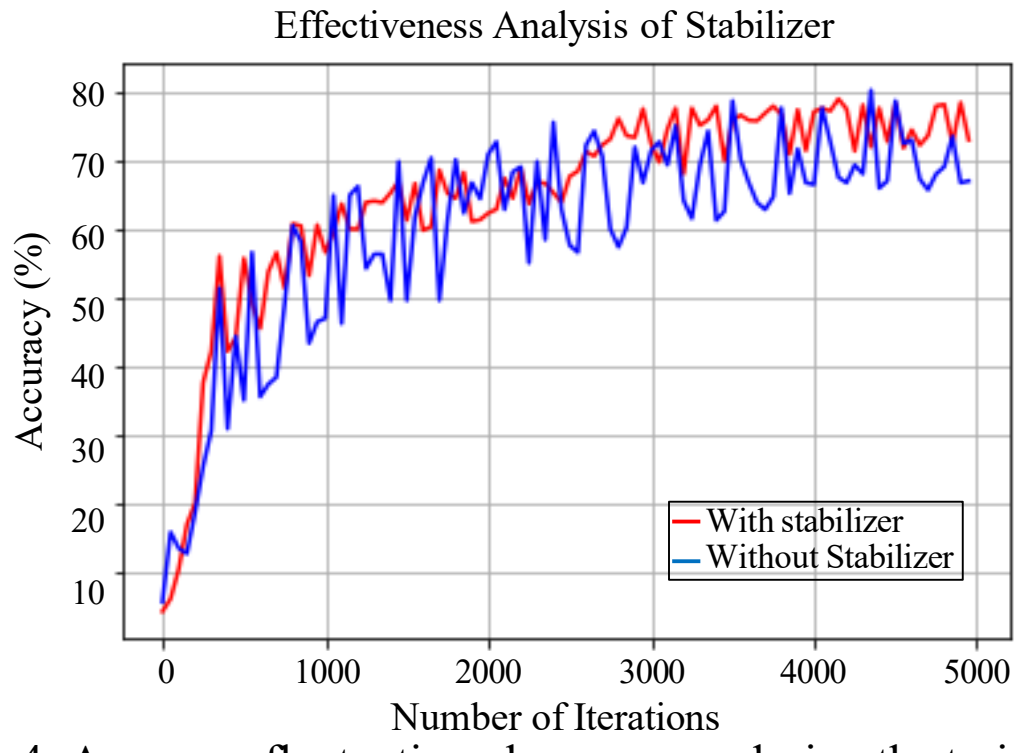

Figure 4. Accuracy fluctuation phenomenon during the training of TPC-ViT. The introduced stabilizer is capable of effectively improving the accuracy and training convergency.

Considering the distribution D as an estimation of how token reduction likelihoods distribute across layers, we regularize D towards a pre-defined prior using Kullback-Leibler (KL) divergence. A new layer-level distribution regularization term $L_{\text{distribution}}$ can be defined as:

$$L_{\text{distribution}} := KL(D \,||\, D^{\text{target-distribution}}), \tag{13}$$

where $D^{\text{target-distribution}}$ indicates a target distribution for the break probability with a guiding stopping layer. We employ a commonly-used Gaussian distribution to define a bell-shaped distribution $D^{\text{target-distribution}}(l)$ as $N(M^{\text{target-depth}}, 1)$ shown in the light blue bell shape of Figure 3, where $M^{\text{target-depth}}$ is the halting depth calculated from Equation (6) for each token. This implicitly encourages the expected sum of break probability value for each token to trigger the stop condition at $M^{\text{target-depth}}$. The final loss function $L_{\text{final}}$ to train the network parameters to control tokens adaptively can be defined as:

$$L_{\text{final}} := L_{\text{task}} + \phi_p L_{\text{ponder}} + \phi_d L_{\text{distribution}}, \tag{14}$$

where a scalar coefficient $\phi_d$ is used to balance the distribution regularization against $L_{\text{distribution}}$ and $L_{\text{task}}$.

### 3.4. Stabilizer

We identify that training our proposed TPC-ViT can show accuracy fluctuations, specially in the early stage of the training procedure (see Figure 4 for an empirical illustration). We posit that this behavior is most likely due to the dense attention map calculation without incorporating any local neighborhood bias. To address this challenge, we propose a stabilizer that introduces the local bias and helps stabilize the process of training and improve the convergence speed. The design philosophy behind the proposed stabilizer is using $\kappa$ similar tokens from the keys for each query to calculate sparse attention maps. Since these similar tokens are from nearby locations, the introduced stabilizer makes TPC naturally inherit the local bias of CNNs without convolutions.

According to [32], the attention matrix computation in a transformer-based model is based on the dot-product. The attention matrix $A \in \mathbb{R}^{n \times n}$ is defined as:

$$A = \text{softmax}\left(\frac{\mathbf{Q}\mathbf{K}^{T}}{\sqrt{d}}\right); \mathbf{Q} = \mathbf{T}\mathbf{W}_q, \mathbf{K} = \mathbf{T}\mathbf{W}_k, \quad (15)$$

where the query matrix $\mathbf{Q} \in \mathbb{R}^{n \times d}$ and key matrix $\mathbf{K} \in \mathbb{R}^{n \times d}$ are generated by the linear projection of the input image token matrix $\mathbf{T} \in \mathbb{R}^{n \times d_m}$ based on the learnable weights matrices $\mathbf{W}_q \in \mathbb{R}^{d_m \times d}$ and $\mathbf{W}_k \in \mathbb{R}^{d_m \times d}$. $n$ indicates the total number of input image tokens. $d$ represents the embedding dimension and $d_m$ denotes the dimension of an input token. The new value matrix $\mathbf{V}_{\text{new}} \in \mathbb{R}^{n \times d}$ can be obtained as:

$$\mathbf{V}_{\text{new}} = A\mathbf{V}; \ \mathbf{V} = \mathbf{T}\mathbf{W}_v, \quad (16)$$

where the value matrix $\mathbf{V} \in \mathbb{R}^{n \times d}$ and $\mathbf{W}_v \in \mathbb{R}^{d_m \times d}$.

Previous works are mainly based on the calculation of all the query-key pairs. In the proposed TPC-ViT, only the top $\kappa$ most similar keys and values for each query are used to compute the attention matrix. For the $i^{\text{th}}$ query, we first compute the Euclidean distance against all the keys, and then obtain its most similar $\kappa$ tokens of keys. These tokens form key-value token pairs $(k_{jl}, v_{jl})$, where $k_{jl} \in S_i^{\text{key}}$, $v_{jl} \in S_i^{\text{value}}$, and $l = 1, \dots, \kappa$. $S_i^{\text{key}}$ denotes a set of the most similar $\kappa$ tokens of keys for the $i^{\text{th}}$ query, and $S_i^{\text{value}}$ indicates a set of the corresponding most similar $\kappa$ tokens of values. Each row of $\hat{A}$ is then derived as:

$$\hat{A}_{i,:} = \text{softmax}\left(\frac{\langle q_i, (k_{j1}, \dots, k_{j\kappa})\rangle}{\sqrt{d}}\right). \quad (17)$$

Then, the value matrix $\mathbf{V}_\kappa \in \mathbb{R}^{n \times d}$ is derived as:

$$\mathbf{V}_\kappa = \hat{A}\hat{\mathbf{V}}_{\text{new}}, \quad (18)$$

where $\hat{A} \in \mathbb{R}^{n \times \kappa}$ and $\hat{\mathbf{V}}_{\text{new}} = (v_{jl})^T \in \mathbb{R}^{\kappa \times d}$. Hence, the dense-dense matrix multiplication in Equation (16) can be turned into the sparse-dense matrix multiplication in Equation (18). This leads to the convergence speed improvement of vision transformer models. We summarize the proposed overall TPC method in **Algorithm 1**. The expanded algorithm can be found in the supplementary.

**Algorithm 1:** Token propagation control (TPC).

**Input:** tokenized input tensor **input** $\in \mathbb{R}^{K \times d}$, $K$, $d$ being number of tokens and embedding dimension; cls indicates classification token; $0 < \delta < 1$
**Output:** aggregated output tensor **out**, ponder loss $\eta$
**Data:** Testing set $x$

**T** = **input** // Tokenized input tensor
**out** = **0** // Output
$\boldsymbol{\eta}$ = **0** // Token ponder loss vector
**cumulation** = **0** // Cumulative break prob.
$\boldsymbol{\zeta}$ = **1** // Token mask vector
**R** = **1** // Remainder
**for** $l = 1, \cdots, L$ **do**
  **t** = Stabilizer(**T**) // Equations 17 to 18
  **t** = $\text{TPC}^{l}_{\text{vit}}(\mathbf{t} \odot \boldsymbol{\zeta})$
  **if** $l < L$ **then**
    $P_1 = \sigma(\gamma \cdot \mathbf{t}_{:,1} + \beta)$ // Pause prob.
    $P_2 = \sigma(\gamma \cdot \mathbf{t}_{:,2} + \beta)$ // Restart prob.
    **b** $= P_1 \cdot P_2$ // Break prob.
  **else**
    **b** $= 1$
  **end if**
  **cumulation** + = **b**
  $\boldsymbol{\eta}$ + = $\boldsymbol{\zeta}$
  Tracker($K$, $\delta$, **R**, **b**, $\boldsymbol{\eta}$) // Equations 6 to 8
  Predict(**out**, $\mathbf{b}_{\text{cls}}$, $\delta$, $\mathbf{cumulation}_{\text{cls}}$, $\mathbf{R}_{\text{cls}}$, $\mathbf{t}_{\text{cls},:}$)
  $\boldsymbol{\zeta} \leftarrow$ **cumulation** $< 1 - \delta$ // Updating mask
**end for**
**return out**, ponder loss $\eta = \frac{\text{sum}(\boldsymbol{\eta})}{K}$

# 4. Experiments

In this section, we first describe the experimental setup in § 4.1. Then, the effectiveness of the proposed TPC-ViT is verified based on comprehensive experiments in § 4.2.

## 4.1. Experimental Settings

The proposed TPC-ViT is evaluated based on the commonly used ImageNet-1K image classification dataset [4] with the input resolution of 224×224. The data-efficient vision transformer architectures (DeiT-{T, S, B}) [31], Swin-T [19] and LV-ViT-{T, S, M} [15] are utilized as the base networks for TPC-ViT. Based on the original training recipe, all models are only trained on the ImageNet-1K dataset without auxiliary images. The default 16× 16 patch resolution is used for DeiT and LV-ViT, and 4× 4 patch resolution is used for Swin. For all experiments, Adam is used for optimization with learning rate $1 \times 10^{-4}$ and cosine learning rate decay. For regularization constants, $\phi_d = 0.1$ and $\phi_p = 5 \times 10^{-4}$ are utilized to scale loss terms. $\gamma = 5$ and $\beta = 10$ are hyperparameters used for sigmoid control gates $P_1(\cdot)$ and $P_2(\cdot)$, shared across all layers. $\xi = 0.5$ is used in the token distribution regularizer. $\delta$ is used as 0.01 and 8 NVIDIA A100 GPUs are used in the training phase. FLOPs calculation follows the work [22]. For throughput, due to the bottleneck of CPU and disk IO, we evaluate the forward pass speed for the derived architectures on 1 A100.

### 4.2. Quantitative and Qualitative Analysis

**State-of-the-art comparison. DeiT**-{T, S, B}[31] firstly introduce convolution-free transformers trained on ImageNet [4] only in less than three days. In Table 2, 3, 4, and 5, the results show that the proposed TPC-ViT performs better than the state-of-the-art methods. Specifically, TPC-ViT outperforms previous state-of-the-art token reduction based methods by 1.1% based on DeiT-S and achieves about 2 × throughput. For DeiT-T architecture, TPC-ViT surpasses previous token reduction related approaches by 2% with much fewer FLOPs. It also achieves the best accuracy with only 70% FLOPs of previous best token reduction method, EViT-B [17], on DeiT-B. **Swin**-T [19] exploits a shifted windowing mechanism to compute *hierarchical* feature representations, which employs MSA in a local window where few tokens are used. Thus, no stabilizer is used in the Swin experiment. TPC-ViT employs 80% FLOPs and maintains the same accuracy of Swin-T in Table 6. **LV-ViT**-{T, S, M} [15] adopt a token labeling training objective that takes advantage of all the image tokens for the training loss calculation in a dense manner. The results in Table 7 demonstrate that the proposed TPC-ViT achieves better accuracy than the previous token reduction method with fewer FLOPs. Although the proposed TPC-ViT exploits fewer number of tokens than DeiT, Swin, and LV-ViT, TPC-ViTs still achieve better accuracy. Note that, existing token reduction methods usually lead to some accuracy drops. This implies that the proposed TPC-ViT is able to effectively distill useful information and discard noises from the input tokens.

**Ablation study for the proposed stabilizer.** To validate the effectiveness of the introduced stabilizer, an ablation study is conducted based on various existing state-of-the-art token reduction approaches. According to Table 8 and Figure 4, the proposed stabilizer is able to improve the model accuracy and stabilize the training process.

**Ablation study for TPC with various $\kappa$ values.** The proposed stabilizer contains a hyper-parameter $\kappa$, which is defined as the top $\kappa$ most similar keys for each query. Empirically, $\kappa$ value of 100 leads to the best accuracy in the experiments. For example, we obtain 72.4% and 73.0% top-1 accuracy using $\kappa$ values of 60 and 100, respectively. Please refer to Table 9 for more details of the ablation study based various $\kappa$ values.

**Ablation study for the introduced token distribution regularizer.** The results in Table 10 demonstrate that the introduced token distribution regularizer is capable of ef-

| Model | Top-1 | FLOPs (G) | Params free |
|---|---|---|---|
| Confidence threshold [18] | 65.8 | 1.1 | Yes |
| PonderNet [1] | 66.2 | 1.0 | Yes |
| DeiT-Ti + PoWER [10] | 69.4 | 0.8 | N/A |
| Similarity gauging [7] | 69.4 | 1.1 | Yes |
| HVT-Ti-1 [22] | 69.6 | 0.6 | N/A |
| DynamicViT-T [23] | 70.9 | 0.9 | No |
| AViT-T [39] | 71.0 | 0.8 | Yes |
| ACT [12] | 71.0 | 1.0 | No |
| DeiT-T [31] | 72.2 | 1.3 | N/A |
| **Ours** | **73.0** | **0.6** | **Yes** |

Table 2. State-of-the-art comparisons based on the ImageNet-1K dataset and DeiT-T.

| Model | Top-1 | Top-5 | Params (M) | FLOPs (G) |
|---|---|---|---|---|
| TokenLearner [24] | 76.1 | - | - | 1.9 |
| HVT-S-1 [22] | 78.0 | 93.8 | 22.09 | 2.4 |
| DeiT-S + PoWER [10] | 78.3 | 94.0 | 22.05 | 2.7 |
| IA-RED$^2$-S [21] | 79.1 | - | - | 3.2 |
| DynamicViT-S [23] | 79.3 | - | 22.8 | 3 |
| Evo-ViT-S [37] | 79.4 | - | - | 3.0 |
| ATS-S [8] | 79.7 | - | 22.05 | 2.9 |
| DGE-S [26] | 79.7 | - | - | 3.1 |
| DeiT-S [31] | 79.8 | 94.9 | 22 | 4.6 |
| **Ours** | **80.8** | **96.1** | **22** | 2.8 |

Table 3. State-of-the-art comparisons based on the ImageNet-1K dataset and DeiT-S. '-' denotes unavailability from previous work.

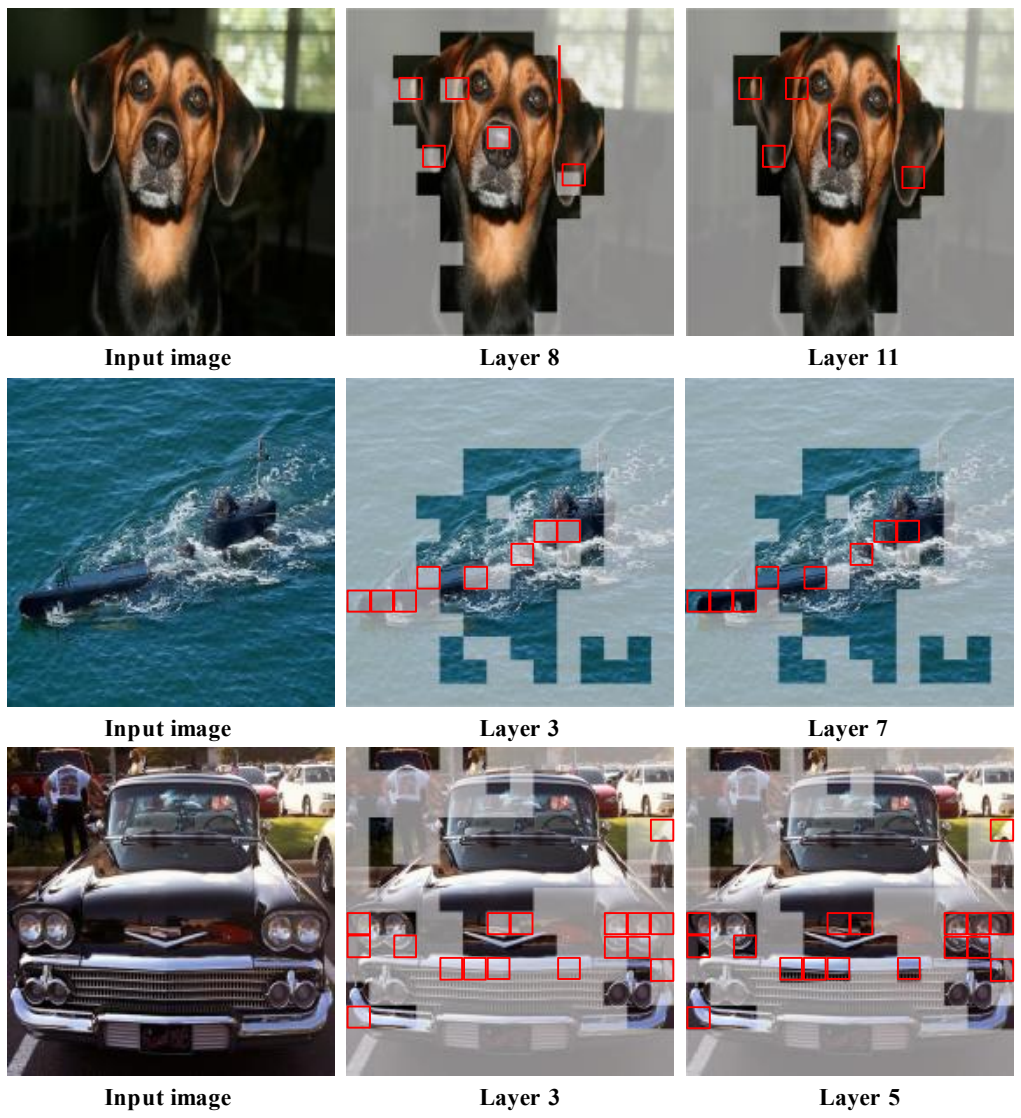

Figure 5. Qualitative results on the ImageNet-1K based on TPC-DeiT-S. The image patches with red bounding boxes show tokens discarded in early layers can be reactivated in the following layers.

| Model | Top-1 Acc | Top-5 Acc | Params (M) | FLOPs (G) | Throughput (img/s) | Image size | $\kappa$ value |
|---|---|---|---|---|---|---|---|
| AViT-S [39] | 78.8 | 93.9 | 22 | 3.6 | 33K | 224x224 | N/A |
| EViT-S [17] | 79.7 | 94.9 | 22.1 | 4 | 31K | 224x224 | N/A |
| DeiT-S [31] | 79.8 | 94.9 | 22 | 4.6 | 26K | 224x224 | N/A |
| **Ours** | **80.8** | **96.1** | **22** | **2.8** | **65K** | **224x224** | **100** |

Table 4. State-of-the-art comparisons, including inference speed, based on the ImageNet-1K dataset and DeiT-S.

| Model | Top-1 | Top-5 | Params (M) | FLOPs (G) |
|---|---|---|---|---|
| IA-RED$^2$-B [21] | 80.3 | - | - | 11.8 |
| DynamicViT-B [23] | 81.3 | - | - | 11.2 |
| Evo-ViT-B [37] | 81.3 | - | - | 10.2 |
| PS-ViT-B [29] | 81.5 | - | - | 9.8 |
| EViT-B [17] | 81.8 | 95.6 | - | 15.3 |
| DeiT-B [31] | 81.8 | 95.6 | 86.6 | 17.6 |
| **Ours** | **81.8** | **95.6** | **86.6** | 10.7 |

Table 5. State-of-the-art comparisons based on the ImageNet-1K dataset and DeiT-B. '-' denotes unavailability from previous work.

| Model | Top-1 | Top-5 | Params (M) | FLOPs (G) |
|---|---|---|---|---|
| Swin-T [19] | 81.2 | 95.5 | 29 | 4.5 |
| **Ours (Swin-T)** | **81.2** | 95.2 | **29** | **3.6** |

Table 6. State-of-the-art comparisons based on the ImageNet-1K dataset and Swin-T.

| Model | Top-1 | Top-5 | Params (M) | FLOPs (G) |
|---|---|---|---|---|
| LV-ViT-T [15] | 79.0 | 94.4 | 8.5 | 2.3 |
| **Ours (LV-ViT-T)** | **79.0** | **94.4** | **8.5** | **1.5** |
| EViT-LV-ViT-S [17] | 83.0 | 96.3 | - | 4.7 |
| LV-ViT-S [15] | 83.2 | 96.3 | 26.3 | 6.6 |
| **Ours (LV-ViT-S)** | **83.2** | 96.2 | **26.3** | **4.3** |
| LV-ViT-M [15] | 83.8 | 96.6 | 55.8 | 16.0 |
| Ours (LV-ViT-M) | 83.7 | **96.6** | **55.8** | **10.5** |

Table 7. State-of-the-art comparisons based on the ImageNet-1K dataset and LV-ViT-*{T, S, M}*. '-' denotes unavailability from previous work.

fectively improving the model accuracy by regularizing the calculation of the break probability, which is probably because the global prior added enhances the learning of TPC.

**Qualitative results.** The qualitative results are shown in Figure 5 to demonstrate the effectiveness of the proposed TPC-ViT. Based on the results, we observe that although the key parts or tokens, *e.g.*, the dog nose and car light, for correctly classifying images are discarded at a shallower layer, they are successfully reactivated to help the model make correct predictions in higher layers.

## 5. Conclusions

In this work, a novel token reduction-based efficient vision transformer, TPC-ViT, has been proposed, which mainly includes three components, *i.e.*, a joint probability-based token propagation controller, a token distribution regularizer, and a model training stabilizer. The proposed controller helps the model adaptively exploit fewer tokens in the attention computation, and the stabilizer helps the model focus on a specific amount of relatively informative tokens for efficient computation. Both of them help the model improve throughput and reduce the FLOPs.

| Model | Top-1 | Top-5 | Params (M) |
|---|---|---|---|
| EViT w/o stab [17] | 79.4 | 94.8 | 22.1 |
| **EViT w/ stab** | **79.6** | **95.1** | **22.1** |
| AViT w/o stab [39] | 78.8 | 93.9 | 22 |
| **AViT w/ stab** | **79.1** | **94.3** | **22** |
| DynamicViT w/o stab [23] | 77.6 | 93.1 | 22.8 |
| **DynamicViT w/ stab** | **77.8** | **93.4** | **22.8** |
| Ours w/o stab | 79.7 | 95.3 | 22 |
| **Ours w/ stab** | **80.1** | **95.7** | **22** |

Table 8. Effectiveness analysis of the introduced stabilizer (stab) based on state-of-the-art token reduction methods. DeiT-S is used in this experiment.

| Model | Top-1 | Top-5 | $\kappa$ value |
|---|---|---|---|
| TPC-ViT w/ stab | 70.6 | 89.7 | 20 |
| TPC-ViT w/ stab | 72.1 | 90.8 | 40 |
| TPC-ViT w/ stab | 72.4 | 91.0 | 60 |
| TPC-ViT w/ stab | 72.7 | 91.2 | 140 |
| TPC-ViT w/ stab | 72.8 | 91.3 | 80 |
| TPC-ViT w/ stab | 72.9 | 91.4 | 120 |
| **TPC-ViT w/ stab** | **73.0** | **91.6** | **100** |

Table 9. Ablation study of the proposed stabilizer (stab) based on various $\kappa$ values. DeiT-T is used in this experiment.

| Model | Top-1 | Top-5 | Params (M) |
|---|---|---|---|
| w/o TDR (TPC-DeiT-S) | 80.1 | 95.7 | 22 |
| **w/ TDR (TPC-DeiT-S)** | **80.8** | **96.1** | **22** |

Table 10. Effectiveness analysis of the proposed token distribution regularizer (TDR). The results show the effectiveness of TDR. TPC-DeiT-S indicates the proposed TPC method based on DeiT-S.